\documentclass[a4paper]{article}
\usepackage{Odyssey2020}
\usepackage{epsfig,amssymb,amsmath,multirow,comment}
\usepackage{amsmath,graphicx,multirow,hyperref,url}

\ninept

\title{Robust Speaker Recognition Using Speech Enhancement And Attention Model}

\name{Yanpei Shi\textsuperscript{*}\thanks{*The first and second author contribute equally to this paper}, Qiang Huang\textsuperscript{*}, Thomas Hain}

\address{Speech and Hearing Research Group\\
Department of Computer Science, University of Sheffield\\
{\small \tt \{YShi30, qiang.huang, t.hain\}@sheffield.ac.uk}}
\begin{document}
\maketitle

\begin{abstract}
In this paper, a novel architecture for speaker recognition 
is proposed by cascading speech enhancement and speaker processing.
It aims to improve speaker recognition performance
when speech signals are corrupted by noise.
Instead of separately processing speech enhancement and speaker recognition, the two
modules are integrated into one framework by a joint optimisation using deep neural networks.
Furthermore, to increase the robustness against noise, a multi-stage attention mechanism is employed to highlight
the speaker related features learned from context information in both time and frequency domains.
To evaluate speaker identification and verification performance of the proposed approach, 
VoxCeleb1, one of mostly used benchmark datasets, is used.
Moreover, the robustness evaluation  
is also conducted on VoxCeleb1 when its being corrupted by three types of interferences,
general noise, music, and babble, at different signal-to-noise ratio (SNR) levels.
The obtained results show that the proposed approach using speech enhancement
and multi-stage attention models outperforms two strong baselines 
in different acoustic conditions in our experiments.


\end{abstract}

\section{Introduction}
The aim of speaker recognition is to recognize speaker identities from their voice characteristics \cite{poddar2017speaker}. 
In recent years, the use of deep learning technologies has
significantly improved speaker recognition performance.
Variani, et al. \cite{variani2014deep} developed the $d$-vector using multiple fully-connected neural network layers, 
and Snyder, et al. \cite{snyder2018x} developed $X$-vectors based on the Time-delayed neural networks (TDNN). 
However, it is still a challenging task when recognizing or verifying speakers in poor acoustic conditions. 

To tackle speech signals corrupted by noise, in this field,
some of previous studies \cite{leglaive2019speech,sadeghi2019audio} tended to recover original signals by removing
noise. Some methods \cite{jang2017enhanced,farahani2006robust} focused on feature extraction from un-corrupted voices,
and some methods \cite{nahma2019adaptive,yao2016priori} tried to estimated speech quality by computing signal-to-noise ratio (SNR).
Although speech enhancement has been used for speaker recognition, 
in most of previous studies it was often processed individually. This might cause
the learned features or enhanced speech signals unable to well meet
the requirement by speaker recognition and verification.
Accordingly, it is highly desirable that both speech enhancement and speaker processing model
can be optimised jointly. 
In \cite{shon2019voiceid}, Shon et.al tried to integrated speech enhancement module and speaker processing
module into one framework.
In this method, the speech enhancement module worked to filter out unnecessary features corrupted by noise
by generating a ratio mask and multiplying element-wise with the original spectrogram for speaker verification.
However, in this work, the speaker verification module was per-trained and frozen when training the speech enhancement network.
This means that the two modules were not optimised jointly.

In addition to joint optimisation, attention mechanisms
have also been widely used for speaker identification and verification \cite{rahman2018attention,zhu2018self,india2019self,an2019deep}.
This is because a neural attention mechanism can
allocate different weights
to different input features. This can hence highlight the relevant information
and reduce the interference caused by irrelevant information. 
In previous studies \cite{wang2018attention, rahman2018attention}, the use of attention models has provided benefits not only
for speech processing, but also for natural language processing (NLP) and image processing 

Wang, et al. \cite{wang2018attention} used an attentive X-vector where a self-attention layer was added 
before a statistics pooling layer to weight each frame. 
Rahman, et al. \cite{rahman2018attention} jointly used attention model and K-max pooling to
selects the most relevant features. 
In \cite{moritz2019triggered}, Moritz, et al. combined CTC (connectionist temporal classification) and attention model 
to improve the performance of end to end speech recognition.
In \cite{mirsamadi2017automatic, zhang2018attention} and  \cite{chorowski2015attention}, 
different attention models were also designed for speech emotion recognition and phoneme recognition, respectively. 
In \cite{bahdanau2014neural}, an attention model was used to allow the each time step 
of decoder to focus on different part of input sentence to search for most relevant words. 
Luong, et al. \cite{luong2015effective} used global attention and local attention, 
where global attention attends to the whole input sentence and local 
attention only looks at a part of the input sentence. Cheng et.al \cite{cheng2016long} proposed self attention 
that relating different positions of the same sentence. 
Woo, et al. \cite{woo2018cbam} used combination of spatial attention and channel attention call CBAM to extract salient features from different dimension of input.

\begin{figure*}[h]
\centering
\includegraphics[height=9cm,width=17cm]{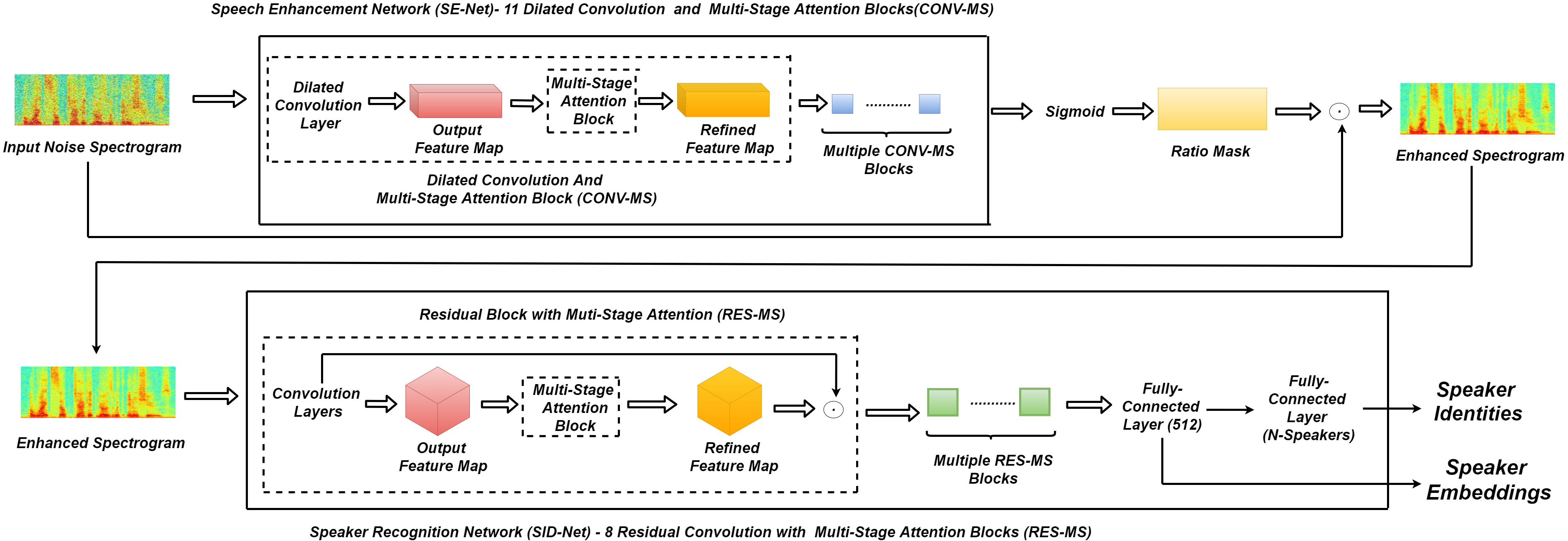}
\caption{Architecture of proposed approach by cascading speech enhancement and speaker recognition. 
SE-Net denotes the speech enhancement network with taking noise spectrogram as input and 
consisting of 11 dilated convolution and multi-stage attention (CONV-MS) blocks. 
SID-Net denotes the speaker recognition network, with taking the enhanced spectrogram as input
and consisting of 8 residual convolution and multi-stage attention (RES-MS) blocks.}
\label{enhance-speaker}
\end{figure*}

To improve the performance for speaker identification and verification, 
and increase the robustness against noise,  
in our proposed approach, the networks of
speech enhancement and speaker
recognition will be cascaded and their parameter
will be optimised jointly by a single loss function.
Simultaneously, 
a multi-stage attention mechanism will be also employed in order
to learn useful features from context information in time, frequency and channel dimensions of the corresponding features. 
The use of multi-stage attention aims to highlight the relevant features to improve the robustness for speaker identification and verification
in noise environment. 
The details of the proposed approach will be depicted in the following sections.


The rest of the paper is organized as follow: 
Section \ref{Model Architecture} presents the cascade structure of our approach and how the attention mechanism
is used in these architecture. The used data set and experiment set-up are introduced in Section \ref{Experiments}. 
The obtained results and related analysis are given in Section \ref{Results}, and finally conclusions are drawn in Section \ref{Conclusion}



\section{Model Architecture}\label{Model Architecture}

\begin{figure*}[h]
\centering
\includegraphics[height=9cm,width=17cm]{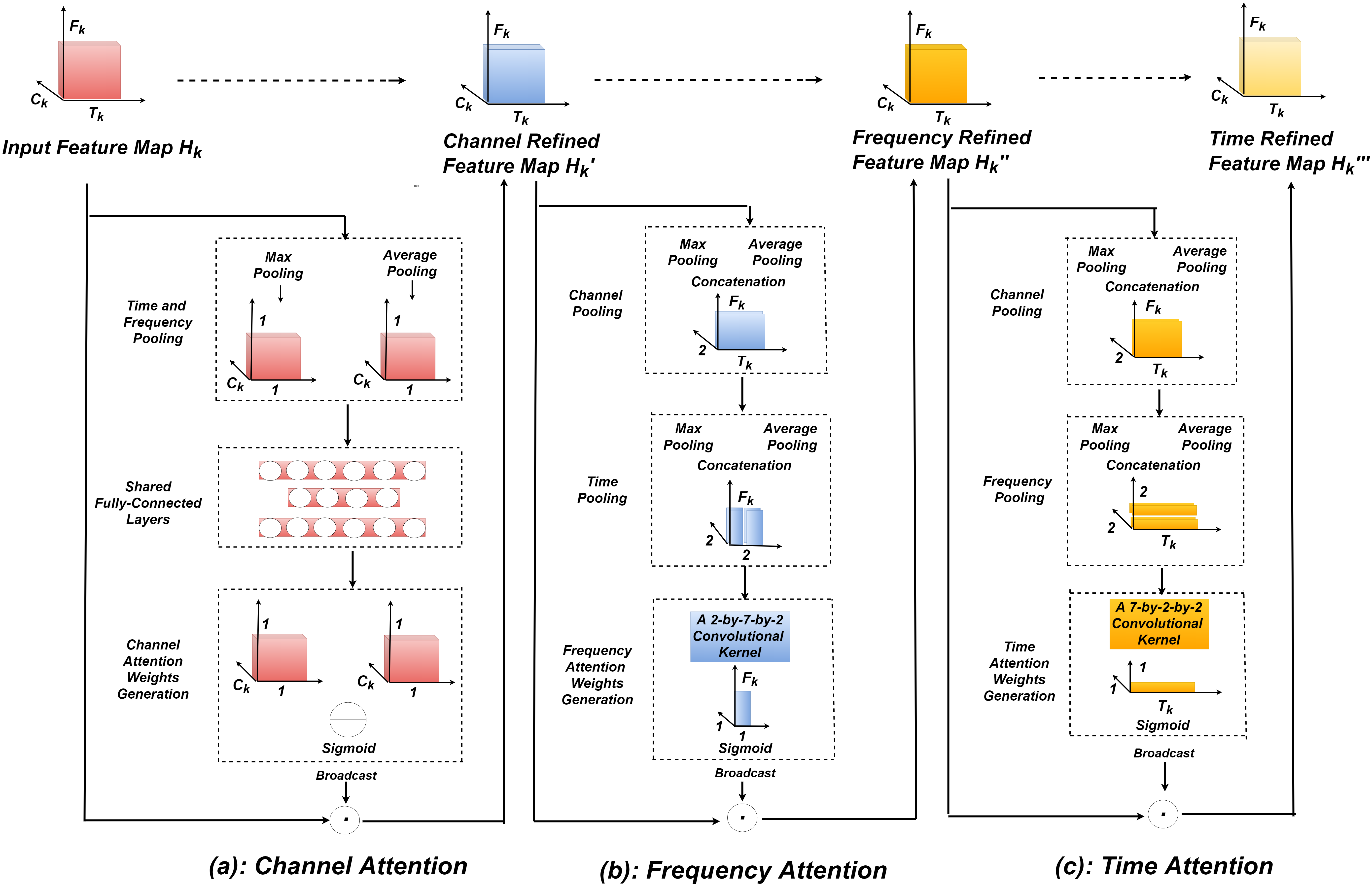}
\caption{The multi-stage (MS) attention consists of three blocks attention block (a): Channel Attention; (b):
Frequency Attention; (c): Time Attention, which are run in a cascading order.}
\label{multi_stage_cnn}
\end{figure*}

\subsection{Speech Enhancement Module}

Figure \ref{enhance-speaker} shows the proposed model architecture consisting of
a speech enhancement module and a speaker recognition module. 
$\boldsymbol X \in  \mathcal {R}^{T \times F \times C})$ represent the input spectrograms,
where $T$, $F$, and $C$ represent the temporal dimension, frequency dimension, and channel dimension, respectively.  
For the input spectrogram, $C$ is set to one, but its value  
is then changed to the number of kernels of a convolutional layer in the propose architecture.

The speech enhancement module consists of multiple Conv-MS blocks, each of them
containing a dilated convolution layer followed by a multi-stage attention block. 
In the attention block, self attention is conducted in time, frequency and channel domains, respectively.

The output of the dilated convolutional layer is denoted as 
$\boldsymbol H_{k} \in \mathcal {R}^{T_{k} \times F_{k} \times C_{k}}$, 
where $k$ means the $k$th CONV-MC block.
The output $\boldsymbol H_{k}'''$ denotes the refined features of the $k$th Conv-MS block,
whose dimension is the same as $\boldsymbol H_{k}$. 

The output of enhancement module is viewed as a ratio mask matrix to
weight the input spectrogram by multiplying it by the corresponding frequency bins and time frames.

\subsection{Speaker Recognition Module}
The speaker recognition module 
consists of multiple residual convolutional blocks and a multi-stage attention block. 
The input of the $k$th residual block is denoted as $\boldsymbol H_{k} \in \mathcal {R}^{T_{k} \times F_{k} \times C_{k}}$, and
the final refined feature map of the $k$th residual block is $\boldsymbol H_{k}'''$. 
Within each residual block, multi-stage is operated sequentially.  
The last residual block is followed by
fully-connected layers, by which the predictions of speaker identities are finally computed using.

\subsection{Multi-Stage Attention (MS)}


Figure \ref{multi_stage_cnn} shows the structure of a multi-stage attention block, which
runs channel attention, frequency attention, and time attention sequentially.
Its mathematics representation can be found in equation \ref{cft}:
\begin{equation}\label{cft}
\footnotesize
\begin{aligned}
\boldsymbol {H^{'}_{k}} &= \boldsymbol {\alpha_{C,k}} \odot \boldsymbol H_{k}\\
\boldsymbol {H^{''}_{k}} &= \boldsymbol {\alpha_{F,k}} \odot \boldsymbol H^{'}_{k}\\
\boldsymbol {H^{'''}_{k}} &= \boldsymbol {\alpha_{T,k}} \odot \boldsymbol {H^{''}_{k}}\\
\end{aligned}
\end{equation}
where $\boldsymbol {\alpha_{C,k}}$, $\boldsymbol {\alpha_{F,k}}$, and $\boldsymbol {\alpha_{T,k}}$ 
represent the implementation of
channel attention, frequency attention and time attention in the $k$th attention block, respectively.

\subsubsection{Channel Attention}

Following the principle of channel attention used in \cite{hu2018squeeze,woo2018cbam},
The working flow of channel attention is 
shown in Figure \ref{multi_stage_cnn} (a) and Equation \ref{channel_att_cnn}. 
\begin{equation}\label{channel_att_cnn}
\footnotesize
\begin{aligned}
\boldsymbol {H^{C}_{k,max}} &=  \boldsymbol {max^{T_{k} \times F_{k} \times 1}(H_{k})}\\
\boldsymbol {H^{C}_{k,avg}} &=  \boldsymbol {avg^{T_{k} \times F_{k} \times 1}(H_{k})}\\
\boldsymbol {S_{max}} &= \mathrm{Relu}\boldsymbol{((H^{C}_{k,max}) {W_{0}}+ {b_{0}}) {W_{1}}}\\
\boldsymbol {S_{avg}} &= \mathrm{Relu}\boldsymbol {((H^{C}_{k,avg}) {W_{0}}+ {b_{0}}) {W_{1}}}\\
\boldsymbol {\alpha^{C,k}} &= \mathrm{Sigmoid}\boldsymbol {({S_{avg}}+{S_{max}})}
\end{aligned}
\end{equation}
where $\boldsymbol {W_{0}} \in \mathcal {R}^{C_{k} \times 100}$, $\boldsymbol {b_{0}} \in \mathcal {R}^{1 \times 100}$ and $\boldsymbol {W_{1}} \in \mathcal {R}^{100 \times C_{k}}$ are the parameters of the $k$th channel attention block. 

In the implementation of channel attention, max pooling and average pooling are firstly 
applied on both time and frequency dimension of $\boldsymbol{H_{k}}$.
Their output $\boldsymbol {H^{C}_{k,avg}} \in \mathcal {R}^{1 \times 1 \times C_{k}}$ 
and $\boldsymbol {H^{C}_{k,max}} \in \mathcal {R}^{1 \times 1 \times C_{k}}$
are then used as the input of two fully connected layers sharing the same parameters and followed by  $Relu$ activation.
The channel attention map $\boldsymbol {\alpha^{C,k}}$ is finally obtained after a Sigmoid activation
is applied to
the summation of $\boldsymbol {S_{avg}}$ and $\boldsymbol {S_{max}}$.
After broadcasting data in $\boldsymbol {\alpha^{C,k}}$ to expand the map size same as $\boldsymbol H_{k}$,
the attention map is multiplied by the original feature map $\boldsymbol H_k$ to generate the refined feature map $\boldsymbol H_{k}'$

\subsubsection{Frequency and Time Attention}


The frequency and time attention block have similar working structure 
when processing their three dimensional input except
that where an attention mechanism is applied to, frequency dimension or time dimension.

\begin{equation}\label{fre_att_cnn}
\footnotesize
\begin{aligned}
\boldsymbol {H^{C'}_{k,max}} &= \boldsymbol{max^{1\times 1\times C_{k}}(H_{k}')}\\
\boldsymbol {H^{C'}_{k,avg}} &= \boldsymbol{avg^{1\times 1\times C_{k}}(H_{k}')}\\
\boldsymbol{H^{C'}_{k,pool}} &= \mathrm{Concat}\boldsymbol {[H^{C'}_{k,avg};H^{C'}_{k,max}]}\\
\boldsymbol {H^{T'}_{k,max}} &= \boldsymbol{max^{T_{k}\times 1\times 1}(H^{C'}_{k,pool})}\\
\boldsymbol {H^{T'}_{k,avg}} &= \boldsymbol{avg^{T_{k}\times 1\times 1}(H^{C'}_{k,pool})}\\
\boldsymbol{H_{k,pool}'} &= \mathrm{Concat}\boldsymbol {[H^{T'}_{k,avg}; H^{T'}_{k,max}]}\\
\boldsymbol {\alpha^{F}_{k}} &= \mathrm{Sigmoid}\boldsymbol {(f^{2 \times 7} (H_{k,pool}'))}
\end{aligned}
\end{equation}

Figure \ref{multi_stage_cnn} (b) shows the working flow of the time attention block
and Equation \ref{fre_att_cnn} shows its implementation in math format.
In the $k$th time attention block, a max pooling and an average pooling are 
firstly applied to channel dimension of the input data $\boldsymbol{H'_{k}}$,
and the corresponding outputs are $\boldsymbol {H^{C}_{k,max}} \in \mathcal {R}^{T_{k} \times F_{k} \times 1} $ 
and $\boldsymbol {H^{C}_{k,avg}} \in \mathcal {R}^{T_{k} \times F_{k} \times 1}$, respectively. 
$\boldsymbol{H^C_{k,pool}} \in \mathcal {R}^{T_{k} \times F_{k} \times 2}$ is obtained by concatenating
the outputs after using poolings. 
On time dimension, the same max pooling and average pooling steps are applied on 
$\boldsymbol{H^C_{k,pool}} \in \mathcal {R}^{T_{k} \times F_{k} \times 2}$  and 
the corresponding outputs are 
$\boldsymbol {H^{T}_{k,avg}} \in \mathcal {R}^{1 \times F_{k} \times 2}$ and 
$\boldsymbol {H^{T}_{k,max}} \in \mathcal {R}^{1 \times F_{k} \times 2}$. Again,
the output after concatenating them on time dimension is  
$\boldsymbol{H_{k,pool}} \in \mathcal {R}^{2 \times F_{k} \times 2}$. 
The frequency attention map $\boldsymbol {\alpha^{F}_{k}}$ is computed 
using a convolution operation with a 2-by-7-by-2 kernel  
followed by a sigmoid activation. The stride value is 1 on frequency dimension during convolution.
The size of $\boldsymbol {\alpha^{F}_{k}}$ is then expanded to the same as 
$\boldsymbol H_{k}''$ by data broadcast. 
The frequency refined feature map $\boldsymbol H_{k}''$ is finally
obtained by the product of $\boldsymbol {\alpha^{F}_{k}}$ and
$\boldsymbol {H'_{k}}$.

\begin{table}[tbh]
\renewcommand{\multirowsetup}{\centering}  
\renewcommand\arraystretch{1.1}
\setlength{\tabcolsep}{6mm}
\centering  
\footnotesize
\begin{tabular}{c|c|c}
\hline
Layer Name& Structure  & Dilation  \\
\hline

\multirow{2}{*}{CONV-MS Block1}
&7x1x48 & 
\multirow{2}{*}{1x1}\\
&MS\\
\hline

\multirow{2}{*}{CONV-MS Block2}
&1x7x48 & 
\multirow{2}{*}{1x1}\\
&MS\\
\hline

\multirow{2}{*}{CONV-MS Block3}
&5x5x48 & 
\multirow{2}{*}{1x1}\\
&MS\\
\hline

\multirow{2}{*}{CONV-MS Block4}
&5x5x48 & 
\multirow{2}{*}{1x2}\\
&MS\\
\hline

\multirow{2}{*}{CONV-MS Block5}
&5x5x48 & 
\multirow{2}{*}{1x4}\\
&MS\\
\hline

\multirow{2}{*}{CONV-MS Block6}
&5x5x48 & 
\multirow{2}{*}{1x8}\\
&MS\\
\hline

\multirow{2}{*}{CONV-MS Block7}
&5x5x48 & 
\multirow{2}{*}{1x1}\\
&MS\\
\hline

\multirow{2}{*}{CONV-MS Block8}
&5x5x48 & 
\multirow{2}{*}{2x2}\\
&MS\\
\hline

\multirow{2}{*}{CONV-MS Block9}
&5x5x48 & 
\multirow{2}{*}{4x4}\\
&MS\\
\hline

\multirow{2}{*}{CONV-MS Block10}
&5x5x48 & 
\multirow{2}{*}{8x8}\\
&MS\\
\hline

\multirow{2}{*}{CONV-MS Block11}
&1x1x1 & 
\multirow{2}{*}{1x1}\\
&MS\\
\hline

\end{tabular}
\caption{Architecture of the speech enhancement network (SE-Net) consists of 11 blocks. 
In each block, a dilated convolutional layer is followed by a multi-stage attention (MS) layer.}
\label{model_sum_se}
\end{table}

\begin{table}[tbh]
\renewcommand{\multirowsetup}{\centering}  
\renewcommand\arraystretch{1.0}
\setlength{\tabcolsep}{6mm}
\centering  
\footnotesize
\begin{tabular}{c|c|c}
\hline
Block Name& Structure  & Output  \\
\hline
\multirow{4}{*}{RES-MS Block1}
&3x3x64 & 
\multirow{4}{*}{150x129}\\
&3x3x64\\&3x3x64\\&MS-ATT\\
\hline

\multirow{4}{*}{RES-MS Block2}
&3x3x128 & 
\multirow{4}{*}{75x65}\\
&3x3x128\\&3x3x128\\&MS-ATT\\
\hline

\multirow{3}{*}{RES-MS Block3}
&3x3x128 & 
\multirow{3}{*}{75x65}\\
&3x3x128\\&MS-ATT\\
\hline

\multirow{4}{*}{RES-MS Block4}
&3x3x256 & 
\multirow{4}{*}{38x33}\\
&3x3x256\\&3x3x128\\&MS-ATT\\
\hline

\multirow{3}{*}{RES-MS Block5}
&3x3x256 & 
\multirow{3}{*}{38x33}\\
&3x3x256\\&MS-ATT\\
\hline

\multirow{3}{*}{RES-MS Block6}
&3x3x256 & 
\multirow{3}{*}{38x33}\\
&3x3x256\\&MS-ATT\\
\hline

\multirow{3}{*}{RES-MS Block7}
&3x3x256 & 
\multirow{3}{*}{38x33}\\
&3x3x256\\&MS-ATT\\
\hline

\multirow{4}{*}{RES-MS Block8}
&3x3x512 & 
\multirow{4}{*}{19x17}\\
&3x3x512\\&3x3x128\\&MS-ATT\\
\hline

Pool&19x1&1x17x512\\
\hline

FC&512& \\
\hline
\end{tabular}
\caption{Architecture of SID-Net consists of 8 blocks. 
Within each block, the multiple convolutional layers are followed by a multi-stage attention (MS) layer before a residual connection.}
\label{sid_sum}
\end{table}

The computation of time attention is similar to frequency attention.
Equation \ref{fre_att_cnn} and Figure \ref{multi_stage_cnn} (c) shows the computation flow. 
The final feature representation is obtained by the multiplication of the previous frequency refined feature map 
and the time attention weights $\boldsymbol {\alpha^{T}_{k}}$.

\begin{equation}\label{time_att_cnn}
\footnotesize
\begin{aligned}
\boldsymbol {H^{C''}_{k,max}} &= \boldsymbol{max^{1\times 1\times C_{k}}(H_{k}'')}\\
\boldsymbol {H^{C''}_{k,avg}} &= \boldsymbol{avg^{1\times 1\times C_{k}}(H_{k}'')}\\
\boldsymbol{H^{C''}_{k,pool}} &= \mathrm{Concat}\boldsymbol {[H^{C''}_{k,avg};H^{C''}_{k,max}]}\\
\boldsymbol {H^{F''}_{k,max}} &= \boldsymbol{max^{1\times F_{k}\times 1}(H^{C''}_{k,pool})}\\
\boldsymbol {H^{F''}_{k,avg}} &= \boldsymbol{avg^{1\times F_{k}\times 1}(H^{C''}_{k,pool})}\\
\boldsymbol{H_{k,pool}''} &= \mathrm{Concat}\boldsymbol {[H^{F''}_{k,avg}; H^{F''}_{k,max}]}\\
\boldsymbol {\alpha^{T}_{k}} &= \mathrm{Sigmoid}\boldsymbol{(f^{7 \times 2} (H_{k,pool}''))}
\end{aligned}
\end{equation}

\section{Experiments}\label{Experiments}

\begin{table}[h]
\renewcommand\arraystretch{1.2}
\setlength{\tabcolsep}{0.65mm}
\centering  
\footnotesize
\begin{tabular}{|c|c|}
\hline
\textbf{Model} & \textbf{Description} \\ \hline
\multirow{2}{*}{\textbf{VoiceID\_loss\cite{shon2019voiceid}}}& baseline done by cascading\\
  & speech enhancement and speaker recognition modules \\\hline
\multirow{2}{*}{\textbf{SID}} & speaker identification baseline using  \\
    & speaker recognition module (SID-Net) only\\  \hline   
\multirow{2}{*}{\textbf{SE+SID}} & joint optimisation of speech enhancement (SE-Net)  \\
      &  and speaker recognition module (SID-Net) \\
      & without using attention mechanism \\ \hline 
\multirow{2}{*}{\textbf{SE-MS+SID}} & proposed model using a joint \\
  & optimisation and a multi-stage attention (MS) \\
  & models in speech enhancement module (SE-Net)\\ \hline
\multirow{2}{*}{\textbf{SE+SID-MS}} & proposed model using a joint \\
  & optimisation and multi-stage attention models \\
  & in speaker recognition module (SID-Net)\\ \hline
\end{tabular}
\caption{Descriptions of five models: three baselines (\textbf{VoiceID\_loss}, \textbf{SID}, and \textbf{SE+SID}) and two proposed approaches (\textbf{SE-MS+SID} and \textbf{SE+SID-MS}).}
\label{tab:models}
\end{table}

\begin{table*}[tbh]
\renewcommand{\multirowsetup}{\centering}  
\renewcommand\arraystretch{1.25}
\setlength{\tabcolsep}{3mm}
\centering  
\footnotesize
\begin{tabular}{c|c|c|c|c|c|c|c|c|c}
\hline
\multirow{2}{*}{\textbf{Noise Type}}& \multirow{2}{*}{\textbf{SNR}} & \multicolumn{2}{c|}{\textbf{SID}}&\multicolumn{2}{c|}{\textbf{SE+SID}} &  \multicolumn{2}{c|}{\textbf{SE-MS
+SID}}&\multicolumn{2}{c}{\textbf{SE+SID-MS}}\\

\cline{3-10}
& & Top1 (\%)& Top5 (\%)&Top1 (\%)& Top5 (\%)& Top1 (\%)& Top5 (\%)&Top1 (\%)& Top5 (\%)\\

\hline
\multirow{5}{*}{\textbf{Noise}}&
0 & 74.1& 86.9& 76.3 & 88.9& \textbf{78.5} & \textbf{90.0} & 77.7&89.2 \\
&5& 79.2& 90.0 & 81.1 &91.8 & \textbf{83.4}& \textbf{92.1}& 81.9&91.8 \\
&10& 83.2& 93.2 & 86.0 &94.7 & \textbf{87.3}& \textbf{95.6}& 86.7&95.1 \\
&15 & 84.9& 94.6& 87.3 &95.8 & \textbf{89.5}& \textbf{96.7}&88.8 &96.0 \\
&20& 87.9& 95.4 & 89.1&96.6 & \textbf{90.9}& \textbf{97.5}& 90.2& 97.0\\
\hline

\multirow{5}{*}{\textbf{Music}}&
0& 65.8& 82.0 & 67.7 &83.7 & \textbf{70.3}& \textbf{84.1}& 69.5& 83.5\\
&5& 76.9& 89.1 & 80.0 &91.0 & \textbf{81.6}& \textbf{91.5}& 80.6 & 90.8\\
&10& 83.8& 93.5 & 85.2 &94.7 & \textbf{86.3}&\textbf{95.3} & 85.8 & 94.7\\
&15& 86.1& 93.9 & 88.4 &95.6 & \textbf{89.1}& \textbf{96.7}& 88.2& 95.4\\
&20& 87.4& 94.7 & 89.1 &96.0 & \textbf{90.2}& \textbf{97.1}& 89.5& 96.6\\
\hline

\multirow{5}{*}{\textbf{Babble}}&
0& 62.4& 80.2 & 65.7 &81.5 & \textbf{67.5}& \textbf{83.0}& 66.6& 81.9\\
&5& 76.2& 87.3 & 78.6 &88.9 & \textbf{80.6}& \textbf{89.9}& 79.3& 89.6\\
&10& 81.4& 92.2 & 84.6 &93.6 & \textbf{86.6}& \textbf{94.5}& 85.3& 83.2\\
&15& 84.0& 92.6 & 86.8 &93.9 & \textbf{88.3}& \textbf{94.7}& 87.6 &94.0 \\
&20& 85.8& 92.9 & 87.1 &94.6 & \textbf{89.0}&\textbf{95.5}& 88.8& 95.2\\
\hline

\textbf{Original}&  & 88.5& 95.9& 89.8 & 96.5 & \textbf{91.9}&\textbf{97.6}& 90.8 & 97.3 \\
\hline
\end{tabular}

\caption{Speaker Identification Results on the Voxcebe1 test data when being corrupted by three types of noise (Noise, Music and Babble) at different SNR (0-20 dB) levels. Four different scenarios are tested: SID-Net (SID), the use of both SE-Net and SID-Net without employing a multi-stage attention (SE+SID), a joint system combing SE-Net with SID-Net, but a multi-stage attention is used only in SE-Net(SE-MS+SID); The SE-Net and SID-Net denotes a joint system, with a multi-stage attention layer being used only in SID-Net(SE+SID-MS).}
\label{identification}
\end{table*}

\begin{table*}[bh]
\renewcommand{\multirowsetup}{\centering}  
\renewcommand\arraystretch{1.3}
\setlength{\tabcolsep}{2mm}
\centering  
\footnotesize
\begin{tabular}{c|c|c|c|c|c|c|c|c|c|c|c}
\hline
\multirow{2}{*}{\textbf{Noise Type}}& \multirow{2}{*}{\textbf{SNR}} & \multicolumn{2}{c|}{\textbf{SID}}& \multicolumn{2}{c|}{\textbf{VoiceID Loss \cite{shon2019voiceid}}} &  \multicolumn{2}{c|}{\textbf{SE+SID}}&\multicolumn{2}{c|}{\textbf{SE-MS+SID}}&\multicolumn{2}{c}{\textbf{SE+SID-MS}}\\

\cline{3-12}
& &EER (\%)& DCF& EER (\%)& DCF& EER (\%)& DCF& EER (\%)& DCF &EER (\%)& DCF\\

\hline
\multirow{5}{*}{\textbf{Noise}}&
0 &16.94 &0.933 & 16.56 & 0.938& 16.20 & 0.912& \textbf{15.95}&\textbf{0.901}& 16.13& 0.908\\
&5 &12.48 &0.855 & 12.26 & 0.830 & 11.99& 0.819& \textbf{11.76}&\textbf{0.805}& 11.78& 0.812\\
&10 &10.03 &0.760& 9.86 & 0.747 & 9.54& 0.732& \textbf{9.17}&\textbf{0.717}& 9.29& 0.727\\
&15 &8.84 &0.648& 8.69 &0.686 & 8.48&  0.665& \textbf{8.08}&\textbf{0.639}& 8.10& 0.641\\
&20 &7.96 &0.594& 7.83& 0.639& 7.52& 0.629& \textbf{7.07}& \textbf{0.615}& 7.09& 0.623\\
\hline

\multirow{5}{*}{\textbf{Music}}&
0 &17.04 &0.940& 16.24 &0.913 & 15.96& 0.901& \textbf{15.58}&\textbf{0.899}& 15.89& 0.904\\
&5 &11.54 &0.828& 11.44&0.818 & 11.15& 0.805& \textbf{10.93}&\textbf{0.791}& 11.04& 0.801\\
&10 &9.69 &0.749& 9.13 &0.733 & 9.12& 0.731& \textbf{8.87}& \textbf{0.714}& 8.97& 0.725\\
&15 &8.40 &0.689& 8.10 &0.677 & 8.08 & 0.643& \textbf{7.62}&\textbf{0.621}& 7.77&0.629\\
&20 &7.70 &0.665& 7.48 &0.635 & 7.39 & 0.619&\textbf{7.13} & \textbf{0.607}& 7.26& 0.614\\
\hline

\multirow{5}{*}{\textbf{Babble}}&
0 &38.90 &1.000& 37.96 &1.000 &37.53 &0.999& 37.55&0.999&  \textbf{37.46} &\textbf{0.998}\\
&5 &28.04 &0.998& 27.12 &0.996 & 26.97&0.979 & 26.42&0.981& \textbf{26.35}& \textbf{0.977}\\
&10 &17.34 &0.917& 16.66 &0.926 & 16.44&0.911 & \textbf{16.30}&\textbf{0.907}& 16.36& 0.911\\
&15 &11.31 &0.795& 11.25 &0.807 & 11.24& 0.801& \textbf{10.89}&\textbf{0.795}& 10.94& 0.801\\
&20 &9.12 &0.720& 8.99 &0.705 & 8.77& 0.695& \textbf{8.39}& \textbf{0.677}& 8.51& 0.688\\
\hline

\textbf{Original}&  &6.92 &0.565 & 6.79  & 0.574 & 6.41 & 0.541 & \textbf{6.18}& \textbf{0.528}& 6.26& 0.535\\
\hline
\end{tabular}
\caption{Speaker Verification Results on Voxceleb1 test data when it being corrupted by different types of noise (Noise, Music and Babble) at different SNR (0-20 dB). Four different scenarios are tested: only use SID-Net (SID); A joint system combining the SE-Net with the SID-Net without a multi-stage attention (SE+SID); A joint system using both SE-Net and SID-Net, but without being used in multi-stage attention (SE-MS+SID); A joint system consisting of SE-Net and SID-Net, with a multi-stage attention being used in SID-Net (SE+SID-MS). The results of VoiceID Loss \cite{shon2019voiceid} is listed and works as a baseline.}
\label{verification}
\end{table*}

\subsection{Data}\label{Data}
In this work, Voxceleb1 \cite{nagrani2017voxceleb} dataset is used to evaluate the proposed approach. 
Voxceleb1 data are extracted from Youtube videos, which
contains 1251 speakers with more than 150 thousand utterances collected "in the wild". 
The average length of the audios in the dataset is 7.8 seconds.

The spectrogram of each recording is used as input features.
Each recording is segmented frames using a 25-ms sliding window with a 10-ms hop, and then
a 512-point FFT is implemented on audio segments. 
In our experiments, a 3-second audio segment are randomly extracted from each recording without any normalization.

To evaluate the robustness of the proposed approach, extra noise from MUSAN dataset is used. 
MUSAN dataset contains three categories of noises: general noise, music and babble \cite{snyder2015musan}. The general noise type contains 6 hours of audio, including DTMF tones, dialtones, fax machine noises et.al. The music type contains 42 hours of music recording from different categories. The babble type contains 60 hours of speech , including read speech from public domain, hearings, committees and debates et.al.

\subsection{Speaker Identification}
In VoxCeleb1 dataset, both training and test sets contain the same number of speakers (1251 speakers) \cite{nagrani2017voxceleb}. 
The training set contains 145265 utterances and the test set contains 8251 utterances. 
In order to reduce possible bias, the MUSAN dataset is also split into two parts for training and test. 
This is to ensure that the noise signals used for training will not be reused for test. 
Each training utterance is mixed with a type of noise at one of five SNR levels.  
For the test set, the same data configuration is set. 
To evaluate the recognition performance, Top-1 and Top-5 accuracy are employed \cite{hajibabaei2018unified}.

\subsection{Speaker Verification}
There contains 148,642 utterances (1211 speakers) in the VoxCeleb1 development dataset,
and 4,874 utterances (40 speakers) in the test dataset \cite{nagrani2017voxceleb}. 
For the speaker verification task, there are total 37,720 test pairs.
The same configuration on the data for speaker recognition task is also set for speaker verification. 
To compare with the baseline introduced in \cite{shon2019voiceid}, the same loss function and similarity measurement (Cosine) 
are used. 
Equal Error Rate (EER) \cite{cheng2004method} and Detection Cost Function (DCF) \cite{van2007introduction} 
are used as evaluation metrics.
DCF is computed as the average of two minimum DCF score (DCF0.01 and DCF0.001) \cite{van2007introduction,xie2019utterance}.

\subsection{Experiment Setup}
To evaluate our proposed approach,
five models including two baselines and three proposed approaches are to be tested 
on the data mentioned in Section 3.2 and 3.3.
As listed in table \ref{tab:models}, 
\textbf{SID} represents the baseline using the SID-Net, and
\textbf{VoiceID\_loss} \cite{shon2019voiceid} represents the baseline done by cascading
speech enhancement and speaker recognition.
\textbf{SE+SID} represents the cascading structure with a joint optimisation with SE-Net and SID-Net.
\textbf{SE-MS+SID} and \textbf{SE+SID-MS} are the two proposed approaches using
multi-stage attention models in either the speech enhancement module (SE-Net) or the speaker recognition module (SID-Net)
besides the joint optimisation used in \textbf{SE+SID}.

\subsection{Network Structure}
Table \ref{model_sum_se} and Table \ref{sid_sum} shows the detailed structure of 
the speech enhancement and speaker recognition module, respectively. 
In the speech enhancement module, 11 dilated convolutional layers are employed.
The speaker recognition module uses the Resnet-20 architecture\cite{he2016deep}, 
due to its effectiveness in speaker recognition \cite{hajibabaei2018unified}. 

For \textbf{SE-MS+SID}, each dilated convolutional layer in the speech enhancement module
is followed by a multi-stage attention module (MS).
For \textbf{SE+SID-MS}, the multi-stage attention module (MS) is inserted into each residual block.  
Each of these two modules are trained independently, and are then fine-tuned by a joint optimisation. 
During training, Adam optimizer \cite{kingmaadam} is used with the initial learning rate
being set to 1e-3 and the decay rate being set to 0.9 for each epoche.

\section{Results}\label{Results}
Table \ref{identification} shows speaker identification results obtained using the models listed in Table \ref{tab:models}. 
Compared to the \textbf{SID} baseline,  
the use of \textbf{SE+SID} yields better performances for speaker identification.
After using multi-stage attention models,  
\textbf{SE+SID-MS} and \textbf{SE-MS+SID}, about 2$\sim$3\% further improvements 
on Top-1 and Top-5 accuracy are obtained
in comparison with the baseline in all noise conditions. 
Compared to \textbf{SE+SID},
the use of attention model can also show about 1$\sim$2\% relative improvement
even if the SNR is at 0dB level.
This case is probably because the use of attention mechanism 
can highlight the speaker related information and reduce the interference 
caused by irrelevant noise signals. 

For the task of speaker verification, table \ref{verification} shows similar
tendencies when implementing all five models on the test data.
It is clear that \textbf{SE+SID} the use of a joint optimisation can performs better
than VoiceID\_loss \cite{shon2019voiceid} using only a pre-trained speaker identification model instead of
a joint optimisation.
In comparison with speaker identification task, the verification improvements obtained using 
\textbf{SE-MS+SID} and \textbf{SE+SID-MS} are relatively smaller.
This is probably because for speaker verification, the similarity between speaker embeddings 
learned from speaker model is computed using Cosine function instead of directly being
computed using the trained speaker recognition model.  
In addition, for speaker verification, the use of attention model in speech enhancement module can yield
slight better performances in almost all conditions except when speeches
are corrupted by Babble noise at 0dB and 5dB SNR levels.
For this case, a possible reason is that the "Babble" noise signals are
relative complication due to its speaker/speech like characteristics.
The use of attention model in speaker recognition module (SID-Net)
might be more suitable to extract speaker relevant information than using an attention model in
the speech enhancement model when acoustic environment is poor.

\begin{figure}[tb]
\centering
\includegraphics[height=6.5cm,width=8.5cm]{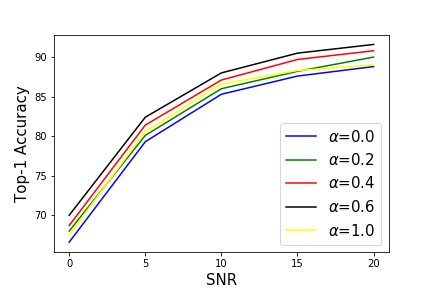}
\caption{The Top-1 Accuracy of the linear combination of the SE-MS+SID and SE+SID-MS results when the noise is "babble". $ \boldsymbol \alpha$ denotes the combination parameter for SE-MS+SID, the combination parameter of SE+SID-MS is $1-  \boldsymbol \alpha$.}
\label{ensumble}
\end{figure}

Figure \ref{ensumble} shows the linear combination of speaker identification results (Top-1 Accuracy) of \textbf{SE-MS+SID} and \textbf{SE+SID-MS}. In the combination results, $\boldsymbol \alpha$ denotes the combination parameter of \textbf{SE-MS+SID} and $1- \boldsymbol \alpha$ denotes that of \textbf{SE+SID-MS}. The two baseline when $\boldsymbol \alpha$ equals to zero and one are also shown. When $\boldsymbol \alpha$ equals to one, the linear combination is the same scenario that only using \textbf{SE-MS+SID}; When $\boldsymbol \alpha$ equals to zero, the linear combination is the same scenario that only using \textbf{SE+SID-MS}. 
It it clear from the figure that when $\boldsymbol alpha$ becomes larger, the final results becomes better. This phenomenon shows the contribution of \textbf{SE-MS+SID} is bigger than \textbf{SE-SID+MS} to the final accuracy. The MS module added in SE module obtains better recognition results. 

\section{Conclusion and Future Work}\label{Conclusion}
In this paper, a joint optimisation by cascading the speech enhancement network
and speaker recognition network was implemented in order to improve speaker identification
and verification performance when speech signals are corrupted by noise. 
Furthermore, a multi-stage attention model also employed
in either the speech enhancement or speaker recognition module
to highlight speaker relevant information.
It is clear that the use of speech enhancement can yield better performances
than the only use of speaker identification model. Moreover, a joint optimisation
and the use of attention model can further increase the robustness of our system
against the interferences caused by different types of noise.

In the future work, the scenario that use multi-stage attention (MS) module in both SE-Net and SID-Net will be tested.
More advanced speech enhancement technologies 
and training strategy such as adversarial training will be studied on other large datasets, such as Voxceleb2. 
Post-processing techniques for speaker embeddings such as PLDA back-end will also be taken into account.

\centerline{\large{\textbf{Acknowledgement}}}
This work was in part supported by Innovate UK Grant number 104264 MAUDIE.

\bibliographystyle{IEEEbib}
\clearpage
\clearpage
\bibliography{Odyssey2020_BibEntries}

%

\end{document}